\documentclass{article}

\usepackage{microtype}
\usepackage{subfigure}
\usepackage{booktabs} 

\usepackage{hyperref}
\usepackage{url}
\usepackage{amsfonts}       
\usepackage{nicefrac}       
\usepackage{microtype}      
\usepackage{natbib}
\usepackage{graphicx}
\usepackage{amsmath}
\usepackage{amssymb}
\usepackage{algorithmic}
\usepackage{enumitem}
\usepackage{verbatimbox}

\newtheorem{theorem}{Theorem}[section]

\usepackage[accepted]{icml2020}

\icmltitlerunning{Towards Stable and Comprehensive Domain Alignment: Max-Margin Domain-Adversarial Training}

\begin{document}

\twocolumn[
\icmltitle{Towards Stable and Comprehensive Domain Alignment: \\Max-Margin Domain-Adversarial Training}



\icmlsetsymbol{equal}{*}

\begin{icmlauthorlist}
\icmlauthor{Jianfei Yang}{ntu}
\icmlauthor{Han Zou}{ucb}
\icmlauthor{Yuxun Zhou}{ucb}
\icmlauthor{Lihua Xie}{ntu}
\end{icmlauthorlist}

\icmlaffiliation{ntu}{Nanyang Technological University, Singapore}

\icmlaffiliation{ucb}{EECS, University of California, Berkeley}

\icmlcorrespondingauthor{Jianfei Yang}{yang0478@e.ntu.edu.sg}

\icmlkeywords{Machine Learning, ICML}

\vskip 0.3in
]



\printAffiliationsAndNotice{} 

\begin{abstract}
  Domain adaptation tackles the problem of transferring knowledge from a label-rich source domain to a label-scarce or even unlabeled target domain. Recently domain-adversarial training (DAT) has shown promising capacity to learn a domain-invariant feature space by reversing the gradient propagation of a domain classifier. However, DAT is still vulnerable in several aspects including (1) training instability due to the overwhelming discriminative ability of the domain classifier in adversarial training, (2) restrictive feature-level alignment, and (3) lack of interpretability or systematic explanation of the learned feature space. In this paper, we propose a novel Max-margin Domain-Adversarial Training (MDAT) by designing an Adversarial Reconstruction Network (ARN). The proposed MDAT stabilizes the gradient reversing in ARN by replacing the domain classifier with a reconstruction network, and in this manner ARN conducts both feature-level and pixel-level domain alignment without involving extra network structures. Furthermore, ARN demonstrates strong robustness to a wide range of hyper-parameters settings, greatly alleviating the task of model selection. Extensive empirical results validate that our approach outperforms other state-of-the-art domain alignment methods. Moreover, reconstructing adapted features reveals the domain-invariant feature space which conforms with our intuition. 
\end{abstract}

\section{Introduction}
Deep neural networks have gained great success on a wide range of tasks such as visual recognition and machine translation~\citep{lecun2015deep}. They usually require a large number of labeled data that can be prohibitively expensive to collect, and even with sufficient supervision their performance can still be poor when being generalized to a new environment. The problem of discrepancy between training and testing data distribution is commonly referred to as \textit{domain shift} or \textit{covariant shift}~\citep{shimodaira2000improving}. To alleviate the effect of such shift, \textit{domain adaptation} sets out to obtain a model trained in a label-rich source domain to generalize well in an unlabeled target domain~\citep{pan2010survey}. Domain adaptation has benefited various applications in many practical scenarios, including but not limited to object detection under challenging conditions~\citep{chen2018domain}, cost-effective learning using only synthetic data to generalize to real-world imagery~\citep{vazquez2013virtual}, etc. 

Prevailing methods for unsupervised domain adaptation (UDA) are mostly based on \textit{domain alignment} which aims to learn domain-invariant features by reducing the distribution discrepancy between the source and target domain using some pre-defined metrics such as maximum mean discrepancy~\citep{gretton2007kernel,gretton2012optimal}. Recently,~\citet{ganin15} proposed to achieve domain alignment by domain-adversarial training (DAT) that reverses the gradients of a domain classifier to maximize domain confusion. Having yielded remarkable performance gain, DAT was employed in many subsequent UDA methods~\citep{long2017conditional, chen2019transferability, liu2019transferable}. 

Nevertheless, DAT with gradient reverse layers still face three critical restrictions when applying it to practical scenarios. (1) DAT cannot continuously provide effective gradients for learning domain-invariant representations. The reason is that the binary domain classifier has high-capacity to discriminate two domains and thus overwhelms adversarial training, which is usually solved by manually adjusting the weights of adversarial loss according to specific tasks such as~\cite{shu2018dirt}. (2) DAT cannot deal with pixel-level \textit{domain shift} that are frequently encountered in visual tasks~\citep{hoffman18a}. (3) The domain-invariant features learned by DAT are only based on intuition and learning theory~\cite{ben2010theory} but difficult to interpret, which impedes the investigation of the underlying mechanism of adversarial domain adaptation.

To overcome the aforementioned difficulties, we propose a novel adversarial approach, namely Max-margin Domain-Adversarial Training (MDAT), to realize stable and comprehensive (\textit{i.e.} both feature-level and pixel-level) domain alignment. MDAT works based on a carefully-designed Adversarial Reconstruction Network (ARN). Specifically, ARN consists of a shared feature extractor, a label predictor, and a reconstruction network (\textit{i.e.} decoder) that serves as a domain classifier. MDAT enables an adversarial game between the feature extractor and the decoder. The decoder focuses on reconstructing features on source domain and pushing target features away from a \textit{margin}, while the feature extractor aims to fool the decoder by generating target features that can be reconstructed. In this adversarial way, three critical issues are subtly solved: (1) the max-margin loss reduces the discriminative capacity of domain classifier, balancing and stabilizing domain-adversarial training; (2) without involving any new network structures, MDAT achieves both pixel-level and feature-level domain alignment; (3) reconstructing adapted features to images reveals how the source and target domains are aligned by adversarial training. 
We evaluate ARN with MDAT on both visual and non-visual UDA benchmarks. It shows more training procedure and achieves significant improvement to DAT on all tasks with pixel-level or higher-level \textit{domain shift}. We also observe that it is insensitive to the choices of hyperparameters and as such is favorable for replication in practice. In principle, our approach is generic and can be used to enhance any domain adaptation methods that leverage domain alignment as an ingredient.

\section{Related Work}
Domain adaptation aims to transfer knowledge from one domain to another. \citet{ben2010theory} provide an upper bound of the test error on the target domain in terms of the source error and the $\mathcal{H}\triangle\mathcal{H}$-distance. As the source error is stationary for a fixed model, the goal of most UDA methods is to minimize the $\mathcal{H}\triangle\mathcal{H}$-distance by reducing some metrics such as Maximum Mean Discrepancy (MMD)~\citep{TzengHZSD14,Longicml15} and CORAL~\citep{sun2016deep}. Inspired by Generative Adversarial Networks (GAN)~\citep{goodfellow2014generative}, \citet{ganin15} proposed to learn domain-invariant features by Domain-Adversarial Training (DAT), which has inspired many UDA methods thereafter. For example, \citet{zhang2019bridging} propose a new divergence for distribution comparison based on minimax optimization and \citet{wang2019negative} discover that filtering our unrelated source samples helps avoid negative transfer in DAT. Adversarial Discriminative Domain Adaptation (ADDA) tried to fool the label classifier by adversarial training but not in an end-to-end manner. CyCADA~\citep{hoffman18a} and PixelDA~\citep{bousmalis2017unsupervised} leveraged GAN to conduct both feature-level and pixel-level domain adaptation, which yields significant improvement yet the network complexity is high. Recent works explore that DAT deteriorates feature learning, and hence they propose to overcome it by generating transferable examples~\citep{liu2019transferable} or involving extra regularizer to retain discriminability~\citep{chen2019transferability}. These approaches can also be directly applied to MDAT for further enhancement.

Another line of approaches that are relevant to our method is reconstruction network (\textit{i.e.} decoder network), which enables unsupervised image-to-image translation by learning pixel-level features~\citep{zhu2017unpaired}. In UDA, \citet{ghifary2016deep} employed a decoder network for pixel-level adaptation, and Domain Separate Network (DSN)~\citep{bousmalis2016domain} further leveraged multiple decoder networks to learn domain-specific features. These approaches treat the decoder network as an independent component for augmented feature learning that is irrelevant to domain alignment~\citep{glorot2011domain}. In this paper, we propose to innovatively utilize decoder network as domain classifier in MDAT which enables both feature-level and pixel-level domain alignment in a stable and straightforward fashion.


\begin{figure*}[!htp]
	\centering
	\includegraphics[width=0.85\textwidth]{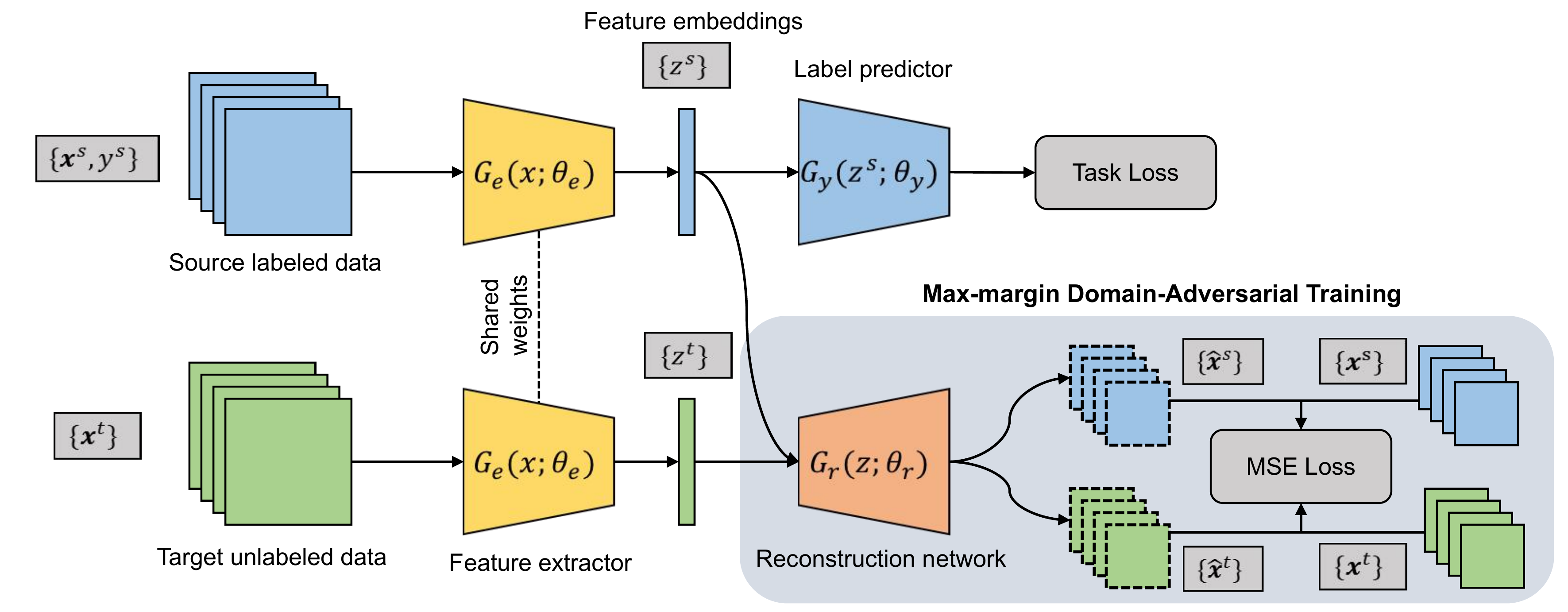}
	\caption{The proposed architecture is composed of a shared feature extractor $G_e$ for two domains, a label predictor $G_y$ and a reconstruction network $G_r$. In addition to the basic supervised learning in the source domain, our adversarial reconstruction training enables the extractor $G_e$ to learn domain-invariant features. Specifically, the network $G_r$ aims to reconstruct the source samples $x^s$ and to impede the reconstruction of the target samples $x^t$, while the extractor $G_e$ tries to fool the reconstruction network in order to reconstruct the target samples $x^t$.}
	\label{fig:framework}
\end{figure*}

\section{Problem Formulation}
\subsection{Problem Definition and Notations}
In unsupervised domain adaptation, we assume that a model works with a labeled dataset $\textbf{X}_S$ and an unlabeled dataset $\textbf{X}_T$. Let $\textbf{X}_S=\{(\textbf{x}^s_i, y^s_i)\}_{i\in [{N_s}]}$ denote the labeled dataset of $N_s$ samples from the source domain, and the certain label $y^s_i$ belongs to the label space $Y$ that is a finite set ($Y={1,2,...,K}$). The other dataset $\textbf{X}_T=\{\textbf{x}^t_i\}_{i\in [{N_t}]}$ has $N_t$ samples from the target domain but has no labels. We further assume that two domains have different distributions, \textit{i.e.} $\textbf{x}^s_i \sim \mathcal{D}_S$ and $\textbf{x}^t_i \sim \mathcal{D}_T$. In other words, there exist some \textit{domain shift}~\citep{ben2010theory} between $\mathcal{D}_S$ and $\mathcal{D}_T$. The ultimate goal is to learn a model that can predict the label $y^t_i$ given the target input $\textbf{x}^t_i$. 

\subsection{Unbalanced Minimax Game in Domain-Adversarial Training}
To achieve domain alignment, Domain-Adversarial Training (DAT) is a minimax game between a shared feature extractor $F$ for two domains and a domain classifier $D$. The domain classifier is trained to determine whether the input sample belongs to the source or the target domain while the feature extractor learns to deceive the domain classifier, which is formulated as:
\begin{equation} 
    \min_{F} \max_{D} \mathbb{E}_{x\sim D_s}[\ln{F(x)}]+\mathbb{E}_{x\sim D_t}[\ln{(1-D(F(x)))}].
\end{equation}
We usually utilize Convolutional Neural Network (CNN) as the feature extractor and fully connected layers (FC) as the domain classifier. Theoretically, DAT reduces the cross-domain discrepancy and helps learn domain-invariant representations~\citep{ganin2016domain}. However, the training of DAT is rather unstable. Without sophisticated tuning of the hyper-parameters, DAT cannot often reach the convergence. Through empirical experiments, we observe that such instability is due to the imbalanced adversarial game between $D$ and $F$. The binary domain discriminator $D$ can easily achieve convergence with very high accuracy at an early training epoch, while it is much harder for the feature extractor $F$ to fool the domain discriminator and to simultaneously perform well on the source domain. In this sense, there is a strong likelihood that the domain classifier overwhelms DAT, and the only solution is to palliate the training of $D$ by tuning the hyper-parameters according to different tasks. In our method, we restrict the capacity of the domain classifier so as to form a minimax game in a harmonious manner. Inspired by the max-margin loss in Support Vector Machine (SVM)~\citep{cristianini2000introduction} (\textit{i.e.} hinge loss), if we push the source domain and the target domain away from a margin rather than as far as possible, then the training task of $F$ to fool $D$ becomes much easier. For a binary domain classifier, we define the margin loss as
\begin{equation}
    \mathcal{L}_{mg}(y)=[0, m-t \cdot y]^{+},
\end{equation}
where $y$ is the predicted domain label, $[\cdot]^{+}$ is $max(0,\cdot)$, $m$ is a positive margin and $t$ is the ground truth label for two domains (assuming $t=-1$ for the source domain and $t=1$ for the target domain). Then we introduce our MDAT scheme based on an innovative network architecture.

\subsection{Max-margin Domain-Adversarial Training}
Besides the training instability issue, DAT also suffers from restrictive feature-level alignment -- lack of pixel-level alignment. To realize stable and comprehensive domain alignment together, we first propose an Adversarial Reconstruction Network (ARN) and then elaborate MDAT.

As depicted in Figure \ref{fig:framework}, our model consists of three parts including a shared feature extractor $G_e$ for both domains, a label predictor $G_y$ and a reconstruction network $G_r$. Let the feature extractor $G_e(\textbf{x};\theta_e)$ be a function parameterized by $\theta_e$ which maps an input sample $\textbf{x}$ to a deep embedding $\textbf{z}$. Let the label predictor $G_y(\textbf{z};\theta_y)$ be a task-specific function parameterized by $\theta_y$ which maps an embedding $\textbf{z}$ to a task-specific prediction $\hat{y}$. The reconstruction network $G_r(\textbf{z};\theta_r)$ is a decoding function parameterized by $\theta_r$ that maps an embedding $\textbf{z}$ to its corresponding reconstruction $\hat{\textbf{x}}$.

The first learning objective for the feature extractor $G_e$ and the label predictor $G_y$ is to perform well in the source domain. For a supervised $K$-way classification problem, it is simply achieved by minimizing the negative log-likelihood of the ground truth class for each sample:
\begin{equation}
    \mathcal{L}_{task}=\sum_{i=1}^{N_s}\mathcal{L}_y(\textbf{x}_i^s, \textbf{y}_i^s)=-\sum_{i=1}^{N_s}\textbf{y}^s_i \cdot \log G_y(G_e(\textbf{x}_i^s;\theta_e);\theta_y),
\end{equation}
where $\textbf{y}^s_i$ is the one-hot encoding of the class label $y^s_i$ and the logarithm operation is conducted on the softmax predictions of the model.

The second objective is to render feature learning to be domain-invariant. This is motivated by the \textit{covariate shift} assumption~\citep{shimodaira2000improving} that indicates if the feature distributions $S(\textbf{z})=\{G_e(\textbf{x};\theta_e)|\textbf{x}\sim \mathcal{D}_S \}$ and $T(\textbf{z})=\{G_e(\textbf{x};\theta_e)|\textbf{x}\sim \mathcal{D}_T \}$ are similar, the source label predictor $G_y$ can achieve a similar accuracy in the target domain. To this end, we first design a decoder network $G_r$ that serves as a domain classifier. In MDAT, we train the decoder network $G_r$ to only reconstruct the features in the source domain and to push the features in the target domain away from a margin. In this way, the decoder has the functionality of distinguishing the source domain from the target domain. The objective of training $G_r$ is formulated as
\begin{equation}\label{eq:r_loss}\small
    \min_{\theta_r} \sum_{i=1}^{N_s+N_t} \mathcal{L}_{mg}(\mathcal{L}_r (\textbf{x}_i))=
    \min_{\theta_r} \sum_{i=1}^{N_s} \mathcal{L}_r (\textbf{x}_i^s) + \sum_{j=1}^{N_t}[m-\mathcal{L}_r(\textbf{x}^t_j)]^+,
\end{equation}
where $m$ is a positive margin and $\mathcal{L}_r(\cdot)$ is the mean squared error (MSE) term for the reconstruction loss defined as
\begin{equation}
    \mathcal{L}_r(\textbf{x}) = || G_r(G_e(\textbf{x};\theta_e);\theta_r) - \textbf{x} ||^2_2,
\end{equation}
where $||\cdot||^2_2$ denotes the squared $L_2$-norm. Compared the normal binary domain classifier (\textit{e.g.} fully connected layers), the decoder network is tailored as a smoothing domain discriminator by separating two domains from a specific margin rather than as far as possible.

Oppositely, to form an adversarial game, the feature extractor $G_e$ learns to deceive $G_r$ such that the learned target features are indistinguishable to the source ones, which is formulated by:
\begin{equation}\label{eq:e_loss}
    \min_{\theta_e} \sum_{j=1}^{N_t} \mathcal{L}_r(\textbf{x}_j^t).
\end{equation}
Then the whole learning procedure of ARN with MDAT can be formulated by:
\begin{align}
    & \min_{\theta_e, \theta_c} \sum_{i=1}^{N_s}\mathcal{L}_y(\textbf{x}_i^s, \textbf{y}_i^s) + \alpha \sum_{j=1}^{N_t} \mathcal{L}_r(\textbf{x}_j^t), \\
    & \min_{\theta_r} \sum_{i=1}^{N_s} \mathcal{L}_r (\textbf{x}_i^s) + \sum_{j=1}^{N_t}[m-\mathcal{L}_r(\textbf{x}^t_j)]^+,
\end{align}
where $\mathcal{L}_y$ denotes the negative log-likelihood of the ground truth class for labeled sample $(\textbf{x}_i^s, \textbf{y}_i^s)$ and $\alpha$ controls the interaction of the loss terms. In the following section, we derive an optimal solution of MDAT and provide theoretical justifications on how MDAT reduces the distribution discrepancy for UDA.


\subsection{Optimal Solution of MDAT}
Considering the adversarial game between a reconstruction network $R$ and a feature extractor $E$ (\textit{i.e.} $G_r$ and $G_e$ in our network, respectively), we prove that if the feature extractor $E$ maps both source domain $x^s\sim P_S(x^s)$ and target domain $x^t\sim P_T(x^t)$ to a common feature space $\mathcal{Z}=\{z=E(x)|z\sim P_z(z) \}$, the MDAT system reaches a Nash Equilibrium. This theoretically explains how MDAT enables the feature extractor to learn domain-invariant features. Similar to EBGAN~\cite{zhao2016energy}, we assume $E$ and $R$ have infinite capacity. Denote $R$ as the MSE of the reconstruction network. We first define two objectives:
\begin{equation}
    V(E,R)=\int_{x^s,x^t}\mathcal{L}_{mg}(x^s,x^t)P_S(x^s)P_T(x^t)dx^s dx^t
\end{equation}
\begin{equation}
    U(E,R)=\int_{x^t}\mathcal{L}_r(x^t)P_T(x^t)dx^t
\end{equation}
In MDAT, we train the feature extractor $E$ to minimize the quantity $V(E,R)$ and train the reconstruction network $R$ to minimize the quantity $U(E,R)$. A Nash equilibrium of our system is a pair $(E^*,R^*)$ that satisfies:
\begin{equation}\label{eq:v}
    V(E^*,R^*) \le V(E^*,R) \quad \forall R 
\end{equation}
\begin{equation}\label{eq:u}
    U(E^*,R^*) \le U(E,R^*) \quad \forall E
\end{equation}

\begin{theorem}
    If a feature extractor $E$ maps both source domain $x^s\sim P_S(x^s)$ and target domain $x^t\sim P_T(x^t)$ to a common feature space $\mathcal{Z}=\{z=E(x)|z\sim P_z(z) \}$, the system reaches a Nash equilibrium $(E^*,R^*)$ and $V(E^*,R^*)=m$.
\end{theorem}
\textit{Proof.} We first prove Eq.\ref{eq:v}:
\begin{align}\nonumber
    V(E^*,R)&=\int_{x^s}P_S(x^s)R(E^*(x^s))dx^s\\ 
    &+\int_{x^t}P_T(x^t)\big[m-R\big(E^*(x^t)\big)\big]^+dx^t\\
    &=\int_z\Big(P_z(z)R(z)+P_z(z)[m-R(z)]^+\Big)dz
\end{align}
As we know $f(x)=x+[m-x]^+$ is monotonically increasing on $[0,+\infty)$, $V(E^*,R)$ reaches its minimum when $R^*(z)=0$:
\begin{align}
    V(E^*,R^*)&=m\int_z P_z(z)dz=m
\end{align}
When $R^*(z)=0$, we expand $U(E,R^*)$ in Eq.\ref{eq:u}:
\begin{align}
    U(E,R^*)&=\int_{x^t}R^*\big(E(x^t)\big)P_T(x^t)dx^t\\
    &=\int_z R^*(z)dz=0
\end{align}
As $U(E,R^*)\geq 0$, we get $U(E^*,R^*) \le U(E,R^*)$.

It can be easily observed that the optimal solution of MDAT is a Nash equilibrium when the feature extractor maps two domains into a common feature space, \textit{i.e.} aligning the distributions in the feature space.

\subsection{Connection to Domain Adaptation Theories}\label{sec:da-theory}
We further investigate how the proposed method connects the learning theory of domain adaptation. The rationale behind domain alignment is motivated from the learning theory of non-conservative domain adaptation problem by Ben-David et al.~\citep{ben2010theory}:

\begin{theorem}
    Let $\mathcal{H}$ be the hypothesis space where $h \in \mathcal{H}$. Let $(\mathcal{D}_S, \epsilon_s)$ and $(\mathcal{D}_T, \epsilon_t)$ be the two domains and their corresponding generalization error functions. The expected error for the target domain is upper bounded by
    \begin{equation}
        \epsilon_t(h) \le \epsilon_s(h) + \frac{1}{2} d_\mathcal{\mathcal{H}\triangle\mathcal{H}}(\mathcal{D}_S,\mathcal{D}_T) + \lambda, \forall h \in \mathcal{H},
    \end{equation}
    where the ideal risk $\lambda=\min_h [\epsilon_s(h)+\epsilon_t(h)]$, and $d_\mathcal{\mathcal{H}\triangle\mathcal{H}}(\mathcal{D}_S,\mathcal{D}_T)=2\sup_{h_1,h_2 \in \mathcal{H}} |\Pr_{x\sim \mathcal{D}_S}[h_1(x)\ne h_2(x)] - \Pr_{x\sim \mathcal{D}_T}[h_1(x)\ne h_2(x)]|$
        
\end{theorem}

Theoretically, when we minimize the $\mathcal{H}\triangle\mathcal{H}$-distance, the upper bound of the expected error for the target domain is reduced accordingly. As derived in DAT~\citep{ganin15}, assuming a family of domain classifiers $\mathcal{H}_d$ to be rich enough to contain the symmetric difference hypothesis set of $\mathcal{H}_p$, such that $\mathcal{H}_p\triangle\mathcal{H}_p=\{ h|h=h_1 \oplus h_2, \, h_1,h_2 \in \mathcal{H}_p \}$ where $\oplus$ is XOR-function, the empirical $\mathcal{H}_p\triangle\mathcal{H}_p$-distance has an upper bound \textit{w.r.t.} the optimal domain classifier $h$:

\begin{equation}\label{eq:bound}\footnotesize
    d_{\mathcal{H}_p \triangle \mathcal{H}_p}(\hat{\mathcal{D}}_S,\hat{\mathcal{D}}_T) \le 2\sup_{h\in \mathcal{H}_d}|\Pr_{\textbf{z}\sim \hat{\mathcal{D}}_S}[h(\textbf{z})=0] + \Pr_{\textbf{z}\sim \hat{\mathcal{D}}_T}[h(\textbf{z})=1]-1|,
\end{equation}
where $\hat{\mathcal{D}}_S$ and $\hat{\mathcal{D}}_T$ denote the distributions of the source and target feature space $\mathcal{Z}_S$ and $\mathcal{Z}_T$, respectively. Note that the MSE of $G_r$ plus a ceiling function is a form of domain classifier $h(\textbf{z})$, \textit{i.e.} $\lceil [m-\mathcal{L}_r(\cdot)]^+-0.5 \rceil$ for $m=1$. It maps source samples to $0$ and target samples to $1$ which is exactly the upper bound in Eq.\ref{eq:bound}. Hence, our reconstruction network $G_r$ maximizes the domain discrepancy with a margin and the feature extractor learns to minimize it adversarially.

\begin{table*}[!htp]
	\small
	\centering
	\caption{We compare with general, statistics-based (\textbf{S}), reconstruction-based (\textbf{R}) and adversarial-based (\textbf{A}) domain adaptation approaches. We repeated each experiment for 3 times and report the average and standard deviation (std) of the test accuracy in the target domain.}
	\label{tab:benchmark}
	\vspace{1mm}
	\begin{tabular}{lcccc}
		\toprule
		\multicolumn{1}{r|}{Source}      & MNIST         & USPS  & SVHN         & SYN \\ 
		\multicolumn{1}{r|}{Target}      & USPS          & MNIST & MNIST          & SVHN  \\ \midrule
		\multicolumn{1}{l|}{\textit{Source-Only model}} & 78.2          & 63.4  & 54.9          & 86.7  \\ 
		\multicolumn{1}{l|}{\textit{Train on target}} & 96.5          & 99.4  & 99.4          & 91.3  \\ \midrule
		\multicolumn{1}{l|}{[\textbf{S}] MMD~\citep{Longicml15}}         & 81.1          & -     & 71.1          & 88.0     \\
		\multicolumn{1}{l|}{[\textbf{S}] CORAL~\citep{sun2016deep}}       & 80.7          & -     & 63.1          & 85.2     \\ \midrule
		\multicolumn{1}{l|}{[\textbf{R}] DRCN~\citep{ghifary2016deep}}    & 91.8          & 73.7  & 82.0          & 87.5   \\
		\multicolumn{1}{l|}{[\textbf{R}] DSN~\citep{bousmalis2016domain}}         & 91.3          & -     & 82.7          & 91.2   \\ \midrule
		\multicolumn{1}{l|}{[\textbf{A}] DANN (DAT)~\citep{ganin2016domain}}        & 85.1          & 73.0  & 74.7          & 90.3   \\
		\multicolumn{1}{l|}{[\textbf{A}] ADDA~\citep{tzeng2017adversarial}}      & 89.4          & 90.1  & 76.0          & -        \\
		\multicolumn{1}{l|}{[\textbf{A}] CDAN~\citep{long2017conditional}}      & 93.9          & 96.9  & 88.5          & -        \\ 
		\multicolumn{1}{l|}{[\textbf{A}] CyCADA~\citep{hoffman18a}}      & 95.6          & 96.5  & 90.4          & -        \\ 
		\multicolumn{1}{l|}{[\textbf{A}] BSP+DANN~\citep{chen2019transferability}  }   & 94.5      & 97.7  & 89.4       & -     \\
		\multicolumn{1}{l|}{[\textbf{A}] MCD~\citep{saito2018maximum}  }   & 96.5      & 94.1  & 96.2       & -     \\ 
		\multicolumn{1}{l|}{[\textbf{A}] CADA~\citep{zou2019consensus}  }   & 96.4      & 97.0  & 90.9       & -     \\  \midrule
		\multicolumn{1}{l|}{\textbf{ARN w.o. MDAT}}       &  93.1$\pm$0.3         &  76.5$\pm$1.2     &    67.4$\pm$0.9      &   86.8$\pm$0.5        \\
		\multicolumn{1}{l|}{\textbf{ARN with MDAT (proposed)}}         & \textbf{98.6$\pm$0.3} & \textbf{98.4$\pm$0.1} & \textbf{97.4$\pm$0.3} & \textbf{92.0$\pm$0.2 }  \\ \bottomrule
	\end{tabular}
\end{table*}

\subsection{Discussions}
Compared with the conventional DAT-based methods that are usually based on a binary logistic network~\citep{ganin15}, the proposed ARN with MDAT is more attractive and incorporates new merits conceptually and theoretically:

\textbf{(1) Effective gradients and balanced adversarial training.} Using the decoder as domain classifier with a margin loss to restrain its overwhelming capacity in adversarial training, the adversarial game can continuously provide effective gradients for training the feature extractor, leading to better alignment and balanced adversarial training. Moreover, through the experiments in Section \ref{sec:exp}, we discover that our method shows more stable training procedure and strong robustness to the hyper-parameters, \textit{i.e.} $\alpha$ and $m$, greatly alleviating the parameters tuning for model selection.

\textbf{(2) Richer information for comprehensive domain alignment.} Rather than typical DAT that uses a bit of domain information, MDAT utilizes the reconstruction network as the domain classifier that captures more domain-specific and pixel-level features during the unsupervised reconstruction~\citep{bousmalis2016domain}. Therefore, MDAT further helps address pixel-level domain shift apart from the feature-level shift, leading to comprehensive domain alignment in a straightforward manner.

\textbf{(3) Feature interpretability for method validation.} MDAT allows us to visualize the features by directly reconstructing target features to images by the decoder network. It is crucial to understand to what extent the features are aligned since this helps to reveal the underlying mechanism of adversarial domain adaptation. We interpret these adapted features in Section \ref{sec:analyses}.

\begin{table}\small
	\centering
	\caption{Comparisons on WiFi gesture recognition.}\label{tab:ida}
	\vspace{1mm}
	\begin{tabular}{l|c}\toprule
		Method & Accuracy (\%)\\\midrule
		\textit{Source-only} & 58.4$\pm$0.7 \\  
		{[\textbf{S}] MMD~\cite{Longicml15}} & 61.2$\pm$0.5 \\ 
		{[\textbf{R}] DRCN~\cite{ghifary2016deep}} & 69.3$\pm$0.3 \\ 
		{[\textbf{A}] DANN (DAT)~\cite{ganin15}} & 68.2$\pm$0.2 \\ 
		{[\textbf{A}] ADDA~\cite{tzeng2017adversarial}} & 71.5$\pm$0.3 \\ 
		{[\textbf{A}] CADA~\cite{zou2019consensus}} & 88.8$\pm$0.1 \\ 
		\textbf{ARN with MDAT} & \textbf{91.3$\pm$0.2} \\
		\bottomrule
	\end{tabular}
\end{table}

\section{Experiment}\label{sec:exp}
We evaluate the proposed approach on several visual and non-visual UDA tasks with varying degrees of \textit{domain shift}. Then detailed analyses are conducted \textit{w.r.t.} toy dataset, parameter sensitivity, gradient and feature visualization. Dataset descriptions and implementation details are attached in the supplementary materials.

\subsection{Setup} 
\textbf{Digits}~\cite{ganin2016domain}. We utilize four digit datasets including \textbf{MNIST}, \textbf{USPS}, \textbf{SVHN} and Synthetic Digits (\textbf{SYN}) that form four transfer tasks: MNIST$\to$USPS, USPS$\to$MNIST, SVHN$\to$MNIST and SYN$\to$SVHN.

\textbf{Office-Home}~\cite{venkateswara2017deep} is a challenging UDA dataset including 15,500 images from 65 categories. It comprises four extremely distinct domains: Artistic images (\textbf{Ar}), ClipArt (\textbf{Cl}), Product images (\textbf{Pr}), and Real-World images (\textbf{Rw}). We evaluate on all twelve transfer tasks.

\textbf{WiFi Gesture Recognition}~\citep{zou2019consensus} consists of six gestures recorded by Channel State Information (CSI)~\citep{xie2018precise}. Each CSI sample is a 2D matrix that depicts the gesture with the surrounding layout environment. Thus, the CSI data collected in two environments forms two domains, which formulates a spatial adaptation problem. 

We compare with state-of-the-art UDA methods that perform three ways of domain alignment. Specifically, \textbf{MMD} regularization~\citep{Longicml15} and \textbf{CORAL}~\citep{sun2016deep} are based on statistical distribution matching. \textbf{DRCN}~\citep{ghifary2016deep} and \textbf{DSN}~\citep{bousmalis2016domain} use the reconstruction network for UDA, while more prevailing UDA methods adopt domain-adversarial training including \textbf{DANN}~\cite{ganin15}, \textbf{ADDA}~\citep{tzeng2017adversarial}, \textbf{CyCADA}~\citep{hoffman18a}, \textbf{CDAN}~\cite{long2017conditional}, \textbf{MCD}~\cite{saito2018maximum}, \textbf{CADA}~\citep{zou2019consensus}, \textbf{TransNorm}~\cite{wang2019transferable} and \textbf{BSP}~\cite{chen2019transferability}. The baseline results are reported from their original papers where available.

We used \textbf{Pytorch} to implement our model. For \textbf{Digits} dataset, we follow the same protocol in~\cite{hoffman18a} and the same network architecture of~\cite{ganin15}. For \textbf{Office-Home}, we adopt ResNet-50 pretrained on ImageNet as our backbone. According to the standard protocols in~\cite{Longicml15}, we employ all the labeled source samples and unlabeled target samples for training. For \textbf{WiFi Gesture Recognition} data, we employ the modified LeNet and the standard protocol in~\citep{zou2019consensus}. The designs of $G_r$ are the inverse of $G_e$ with pooling operation replaced by upsampling. We fix $\alpha=0.02$ and $m=5$ in all the experiments, which are obtained on \textbf{SVHN}$\to$\textbf{MNIST} by Baysian optimization~\citep{malkomes2016bayesian}. We adopt mini-batch SGD optimizer with momentum of 0.9 and the progressive training strategy in DANN~\citep{ganin15}.


\subsection{Overall Results}
The classification accuracies on Digits are shown in Table~\ref{tab:benchmark}. Our method outperforms all other methods on four transfer tasks. Specifically, for \textbf{SVHN$\to$MNIST} where severe pixel-level domain shift exists, our method significantly improves \textbf{DANN} by 22.7\%, which justifies the efficacy of \textbf{ARN} for addressing pixel-level shift. Our method also performs well when the target domain are quite small, achieving 98.6\% accuracy on \textbf{MNIST$\to$USPS}. In Table~\ref{tab:ida}, our method improves the source-only model by 32.9\% on WiFi spatial adaptation problem, which indicates that \textbf{MDAT} is also helpful for non-visual domain adaptation problem. Table~\ref{tab:office-home} shows the performance on large-scale dataset and \textbf{MDAT} yields better performance against other domain alignment approaches.

\begin{table*}[htbp]
      \centering
      \caption{Accuracy (mean) of unsupervised domain adaptation on \textit{Office-Home} datasets across 3 independent runs.}\label{tab:office-home}
      \resizebox{\textwidth}{!}{
      \begin{tabular}{l|cccccccccccc|c}
         \toprule
         Method    & Ar$\to$Cl         & Ar$\to$Pr       & Ar$\to$Rw         & Cl$\to$Ar       & Cl$\to$Pr         & Cl$\to$Rw         & Pr$\to$Ar         & Pr$\to$Cl         & Pr$\to$Rw         & Rw$\to$Ar         & Rw$\to$Cl         & Rw$\to$Pr         & Avg           \\ \midrule
        ResNet-50~\cite{he2016deep}        & 34.9          & 50.0          & 58.0          & 37.4          & 41.9          & 46.2          & 38.5          & 31.2          & 60.4          & 53.9          & 41.2          & 59.9          & 46.1          \\
        DAN~\cite{Longicml15}              & 43.6          & 57.0          & 67.9          & 45.8          & 56.5          & 60.4          & 44.0          & 43.6          & 67.7          & 63.1          & 51.5          & 74.3          & 56.3          \\
        DANN (DAT)~\cite{ganin15}             & 45.6          & 59.3          & 70.1          & 47.0          & 58.5          & 60.9          & 46.1          & 43.7          & 68.5          & 63.2          & 51.8          & 76.8          & 57.6          \\
        TransNorm+DANN~\cite{wang2019transferable}             & 43.5          & 60.9          & 72.1          & 51.0          & 61.5          & 62.5          & 49.6          & 46.8          & 70.4          & 63.7          & 52.2          & 77.9          & 59.3          \\
        CDAN~\cite{long2017conditional}             & 49.0          & 69.3          & 74.5          & 54.4          & 66.0          & 68.4          & 55.6          & 48.3          & \textbf{75.9}          & 68.4          & \textbf{55.4}          & 80.5          & 63.8          \\
        \textbf{ARN with MDAT} (Proposed) & \textbf{51.3} & \textbf{69.7} & \textbf{76.2} & \textbf{59.5} & \textbf{68.3} & \textbf{70.0} & \textbf{57.2} & \textbf{48.9} & 75.8 & \textbf{69.1} & 55.3 & \textbf{80.6} & \textbf{65.2} \\  \bottomrule
      \end{tabular}
  }
\end{table*}

\begin{figure*}[!htp]
	\centering
	\subfigure[Source-only Model]{
		\includegraphics[width=0.26\textwidth]{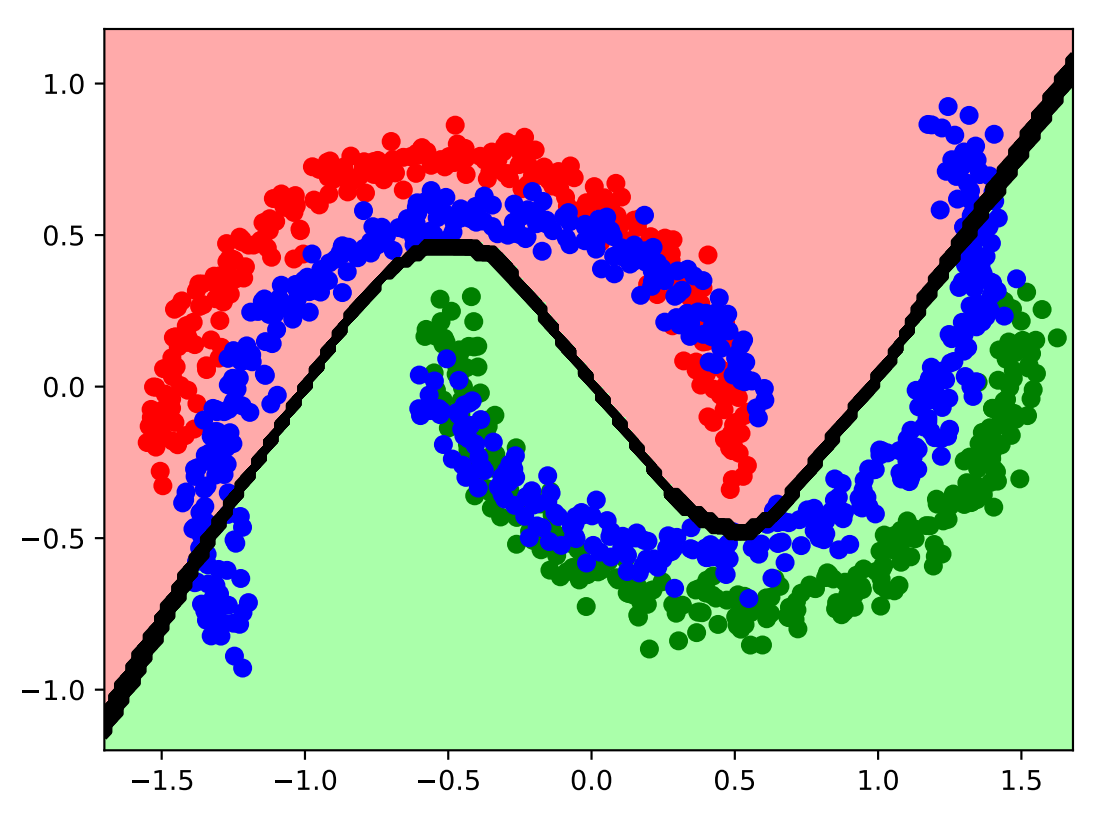}\label{fig:toy-source}}
	\subfigure[DANN]{
		\includegraphics[width=0.26\textwidth]{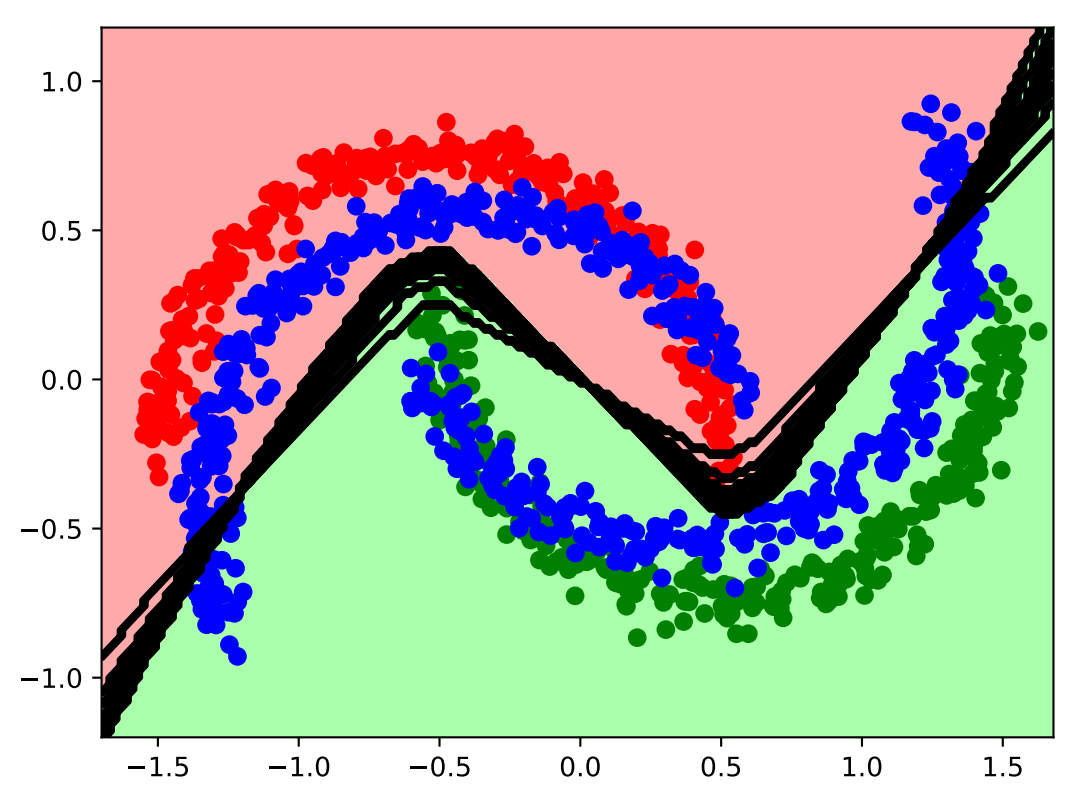}\label{fig:toy-dann}}
	\subfigure[MDAT]{
		\includegraphics[width=0.26\textwidth]{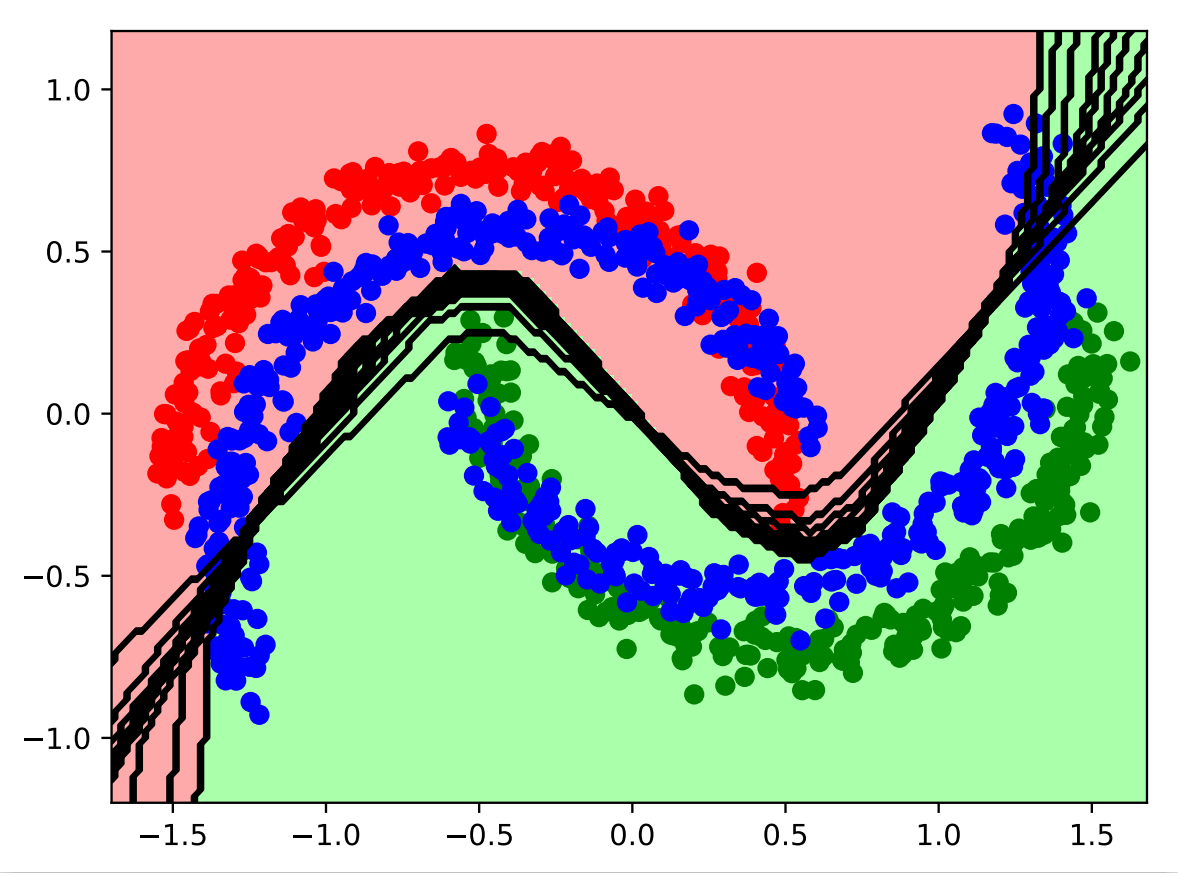}\label{fig:toy-mdat}}
	\caption{The \textit{inter-twining moons} toy problem. Red and green dots indicate source samples, while blue dots are target samples. Black lines indicate the changes of decision boundaries during 10 epochs training.}
	\label{fig:toydata}
\end{figure*}

\begin{figure}[!htp]
	\centering  
	\subfigure[Convergence]{
		\label{fig:loss-compare}
		\includegraphics[width=0.23\textwidth]{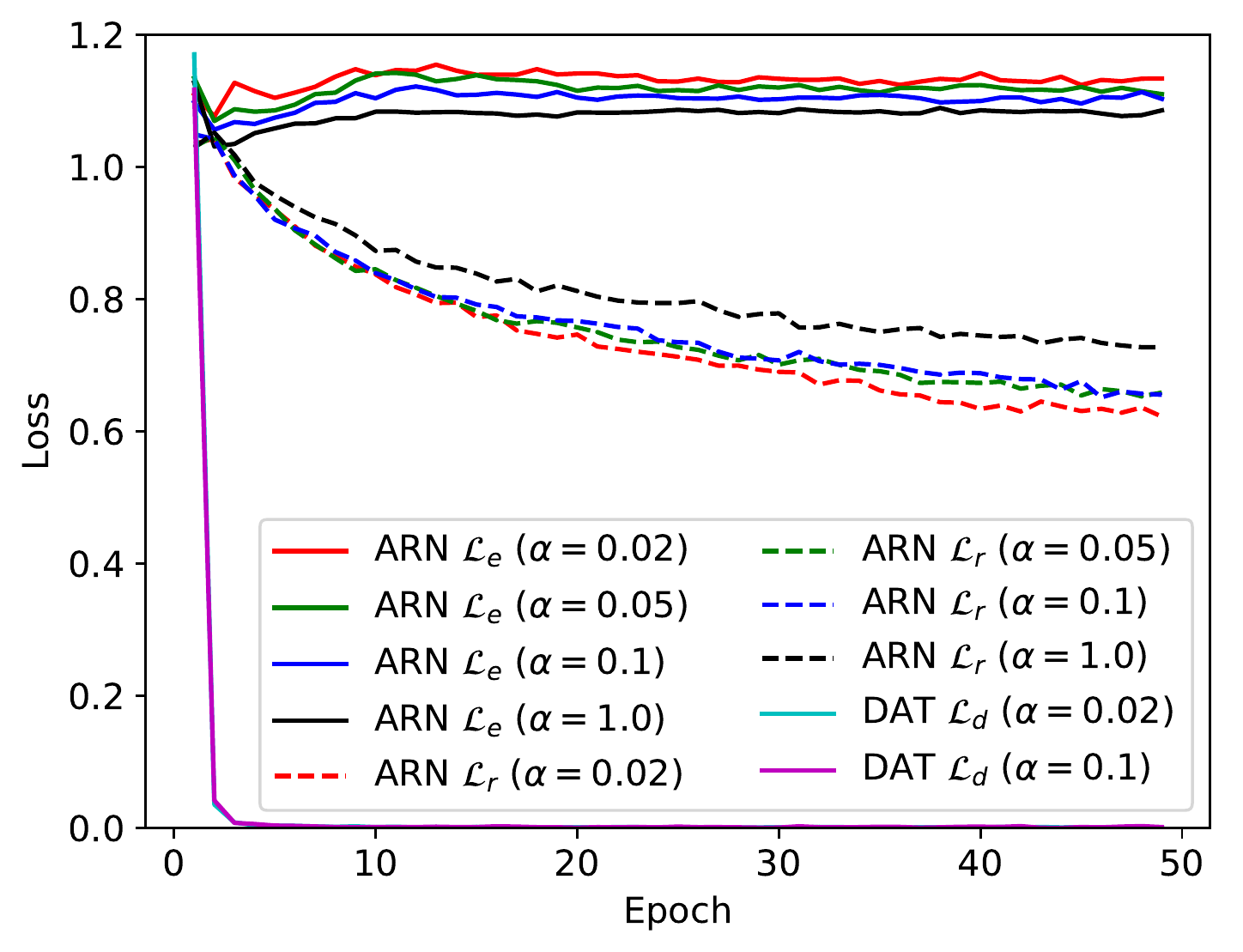}\label{fig:train-loss}}
	\subfigure[Test Accuracy]{
		\label{fig:acc-compare}
		\includegraphics[width=0.23\textwidth]{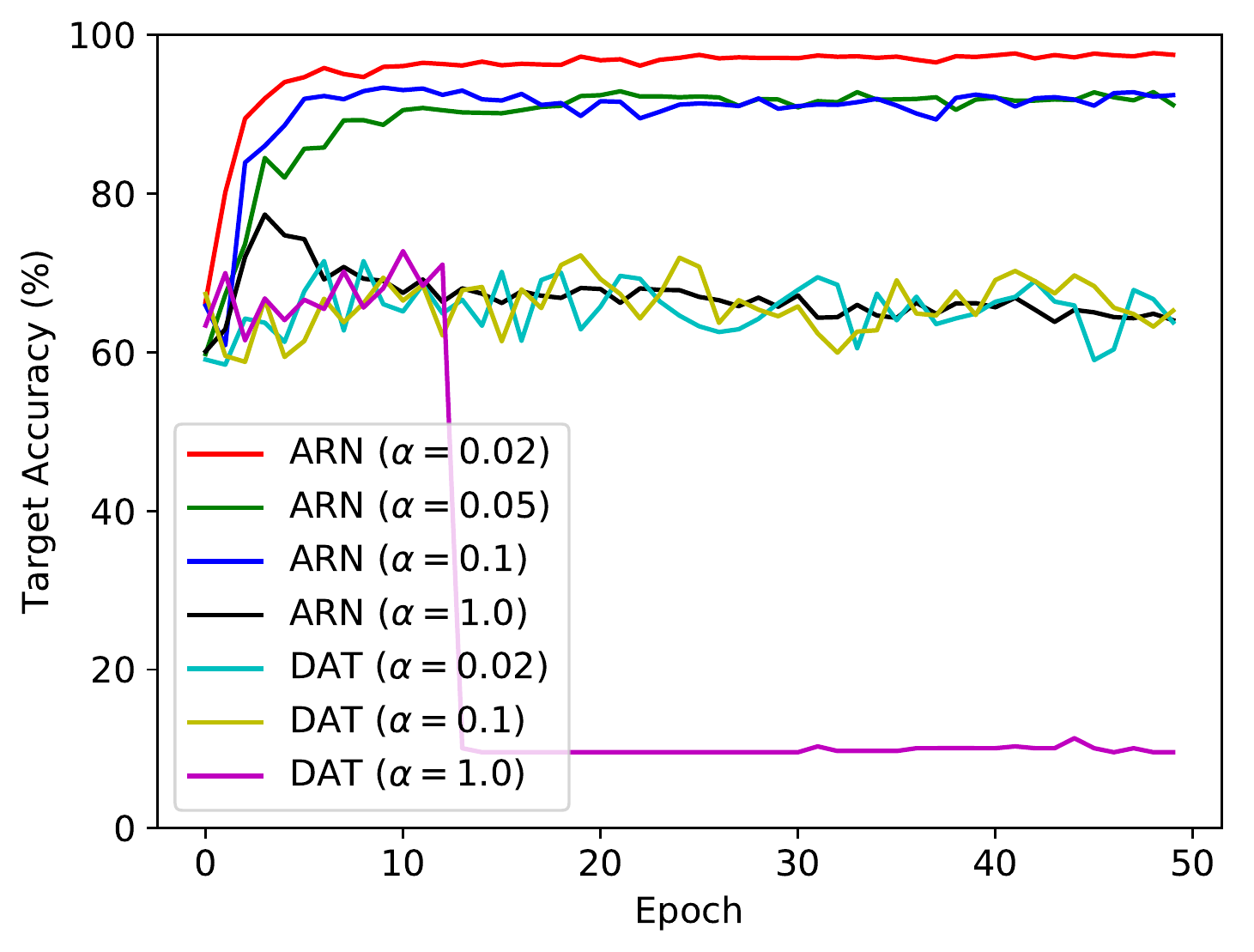}}
	\caption{The training procedure \textit{w.r.t.} loss and test accuracy. $\mathcal{L}_e$ is the training loss of reconstructing target samples in Eq. \ref{eq:e_loss}. $\mathcal{L}_r$ is the training loss of reconstruction network in Eq. \ref{eq:r_loss}. $\mathcal{L}_d$ is the domain loss in DAT~\citep{ganin15}. $\alpha$ is the penalty term of $\mathcal{L}_e$ and $\mathcal{L}_d$ in MDAT and DAT, respectively.}
	\label{fig:training}
\end{figure}

\subsection{Analyses}\label{sec:analyses}
\textbf{Ablation study.} To verify the contribution of the reconstruction network $G_r$ and \textbf{MDAT}, we discard the term $\mathcal{L}_r(\textbf{x}^t)$ in Eq.\ref{eq:r_loss}, and evaluate the method, denoted as \textbf{ARN w.o. MDAT} in Table \ref{tab:benchmark}. Comparing it with source-only model, we can infer the improvement of reconstructing target samples. \textbf{ARN w.o. MDAT} improves tasks with low-level \textit{domain shift} such as \textbf{MNIST$\leftrightarrow$USPS}, which conforms with our discussion that unsupervised reconstruction is instrumental in learning pixel-level features. Comparing \textbf{ARN w.o. MDAT} with the original \textbf{ARN}, we can infer the contribution of \textbf{MDAT}. Table \ref{tab:benchmark} shows that the \textbf{MDAT} achieves an impressive margin-of-improvement. For \textbf{USPS$\to$MNIST} and \textbf{SVHN$\to$MNIST}, the MDAT improves \textbf{ARN w.o. MDAT} by around 30\%. It demonstrates that MDAT that helps learn domain-invariant features is the main reason for the tremendous improvement.


\textbf{Toy dataset.} We study the behavior of MDAT on a variant of \textit{inter-twinning moons} 2D problem, where the target samples are rotated $30^{\circ}$ from the source samples. 300 samples are generated for each domain using \textbf{scikit-learn}~\cite{pedregosa2011scikit}. The adaptation ability is investigated by comparing MDAT with DANN and source-only model. As shown in Figure~\ref{fig:toydata}, we visualize the changing boundaries during 10 epochs training. In Figure~\ref{fig:toy-source}, the model is overfitting the source domain, and the decision boundary does not change. In Figure~\ref{fig:toy-dann} and~\ref{fig:toy-mdat}, both DANN and MDAT adapt the boundaries to the target samples, but MDAT shows faster and better adaptation during 10 epochs. Integrating the training procedure of \textbf{SVHN}$\to$\textbf{MNIST} in Figure~\ref{fig:train-loss}, we justify that more effective gradients are provided by MDAT for better adaptation performance.

\begin{table}[!t]
	\centering
	\caption{Accuracy (\%) on \textbf{SVHN$\to$MNIST}.}\label{tab:sensitivity}
	\vspace{1mm}
	\scalebox{0.72}{
		\begin{tabular}{c|cccccccc}
			\toprule
			$\alpha$ & 0.01 & 0.03 &  0.07 & 0.1 & 0.2 & 0.3 & 0.5 & 1.0 \\ \hline
			DANN & 71.1 & 74.1 & 72.7 & 74.1 & 74.7 & 9.6 & 9.7 & 10.3\\
			ARN ($m=1$) &   95.7  &  95.9  &  93.3   &  93.2  &  80.1 & 75.3 & 73.1  &  67.5 \\ \midrule
			$m$ & 0 & 0.1 & 0.3 & 0.5 & 1.0 & 2.0 & 5.0 & 10.0 \\ \hline
			ARN ($\alpha=2e^{-2}$)  & 64.3 & 64.5  &  75.2  &  90.0  & 96.0  &  97.4 & 97.7 & 96.7 \\ \bottomrule
	\end{tabular}}
\end{table}  

\textbf{Gradients and stability analysis.} We further study the training procedure of MDAT on \textbf{SVHN$\to$MNIST} \textit{w.r.t.} loss and target accuracy in Figure~\ref{fig:loss-compare} and \ref{fig:acc-compare}, respectively. In Figure \ref{fig:loss-compare}, ARN has steadily decreasing loss ($\mathcal{L}_r$) for all $\alpha$, but the domain loss in DAT ($\mathcal{L}_d$) becomes extremely small at the beginning. These observations conform with our intuition: the domain classifier in DAT is too strong to impede the adversarial training, while MDAT provides more effective gradients for training feature extractor by restricting the capacity of domain classifier. With effective gradients, the adversarial game is more balanced, which is validated in Figure \ref{fig:acc-compare} where the test accuracy of ARN is more stable than that of DAT across training epochs.

\begin{table*}[t]
	\centering
	\caption{Visualizing the source image, target images and reconstructed target images (R-Target Images) for four digit adaptation tasks.}\label{tab:visual}
	\vspace{1mm}
	\scalebox{0.9}{
		\begin{tabular}{|c|c|c|c|}\toprule
			& Source Images        & Target Images        & R-Target Images \\ \midrule
			\textbf{MNIST$\to$USPS}  &        \includegraphics[width=0.22\textwidth]{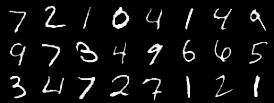}  &       \includegraphics[width=0.22\textwidth]{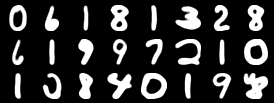}               &       \includegraphics[width=0.22\textwidth]{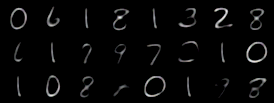}           \\ \midrule
			\textbf{USPS$\to$MNIST}  &          \includegraphics[width=0.22\textwidth]{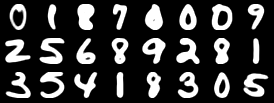}            &          \includegraphics[width=0.22\textwidth]{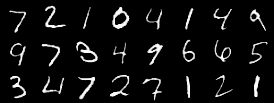}            &             \includegraphics[width=0.22\textwidth]{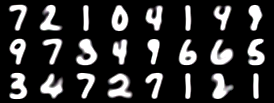}              \\ \midrule
			\textbf{SVHN$\to$MNIST}  &           \includegraphics[width=0.22\textwidth]{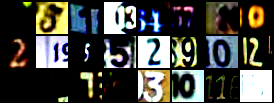}           &            \includegraphics[width=0.22\textwidth]{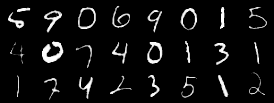}          &           \includegraphics[width=0.22\textwidth]{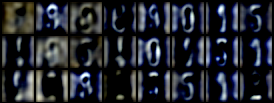}                \\ \midrule
			\textbf{SYN$\to$SVHN}  &            \includegraphics[width=0.22\textwidth]{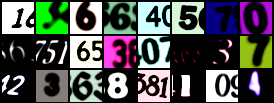}          &           \includegraphics[width=0.22\textwidth]{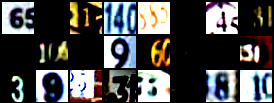}           &             \includegraphics[width=0.22\textwidth]{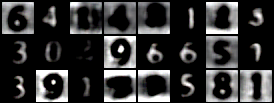}
			\\ \bottomrule
	\end{tabular}}
\end{table*}

\begin{figure*}[t]
	\centering  
	\subfigure[Source-only Model]{
		\label{fig:source-tsne}
		\includegraphics[width=0.3\textwidth]{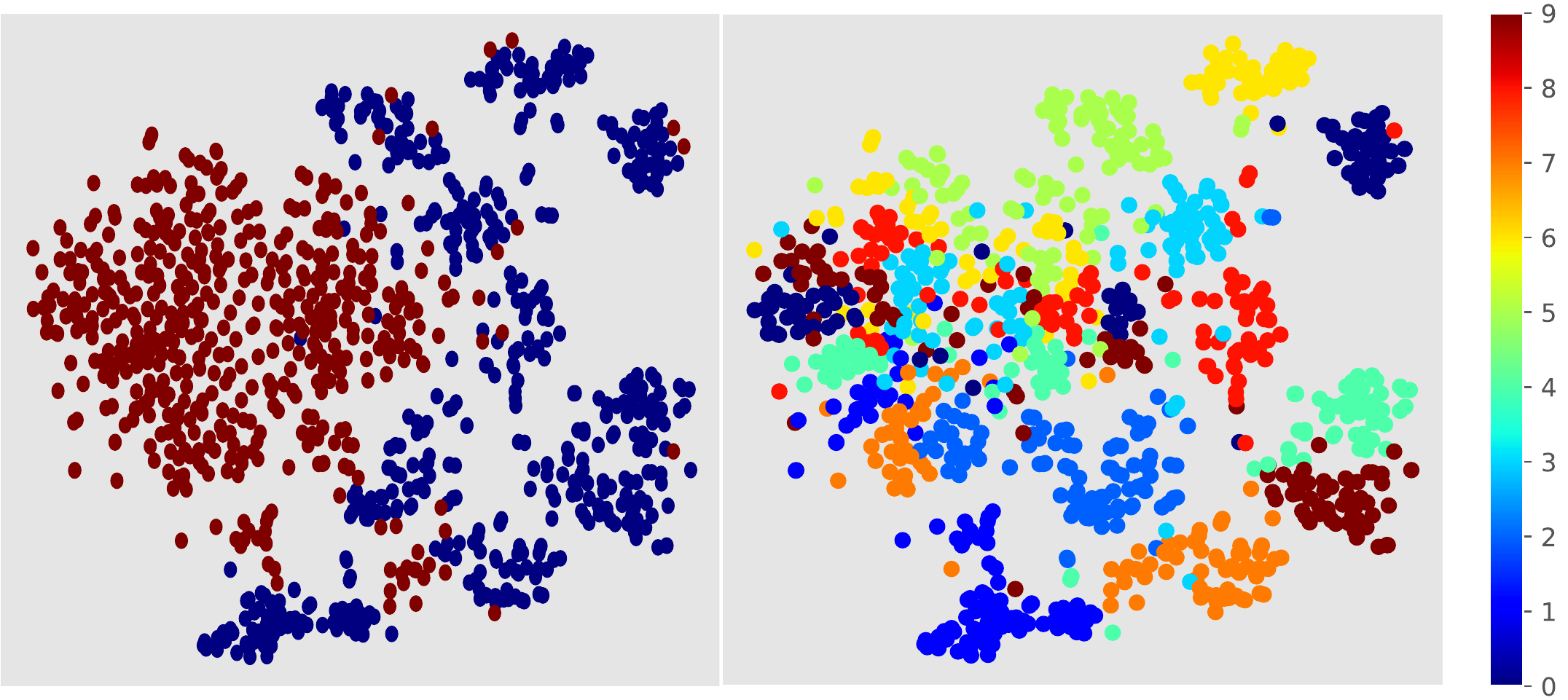}}
	\subfigure[DANN]{
		\label{fig:dann2-tsne}
		\includegraphics[width=0.3\textwidth]{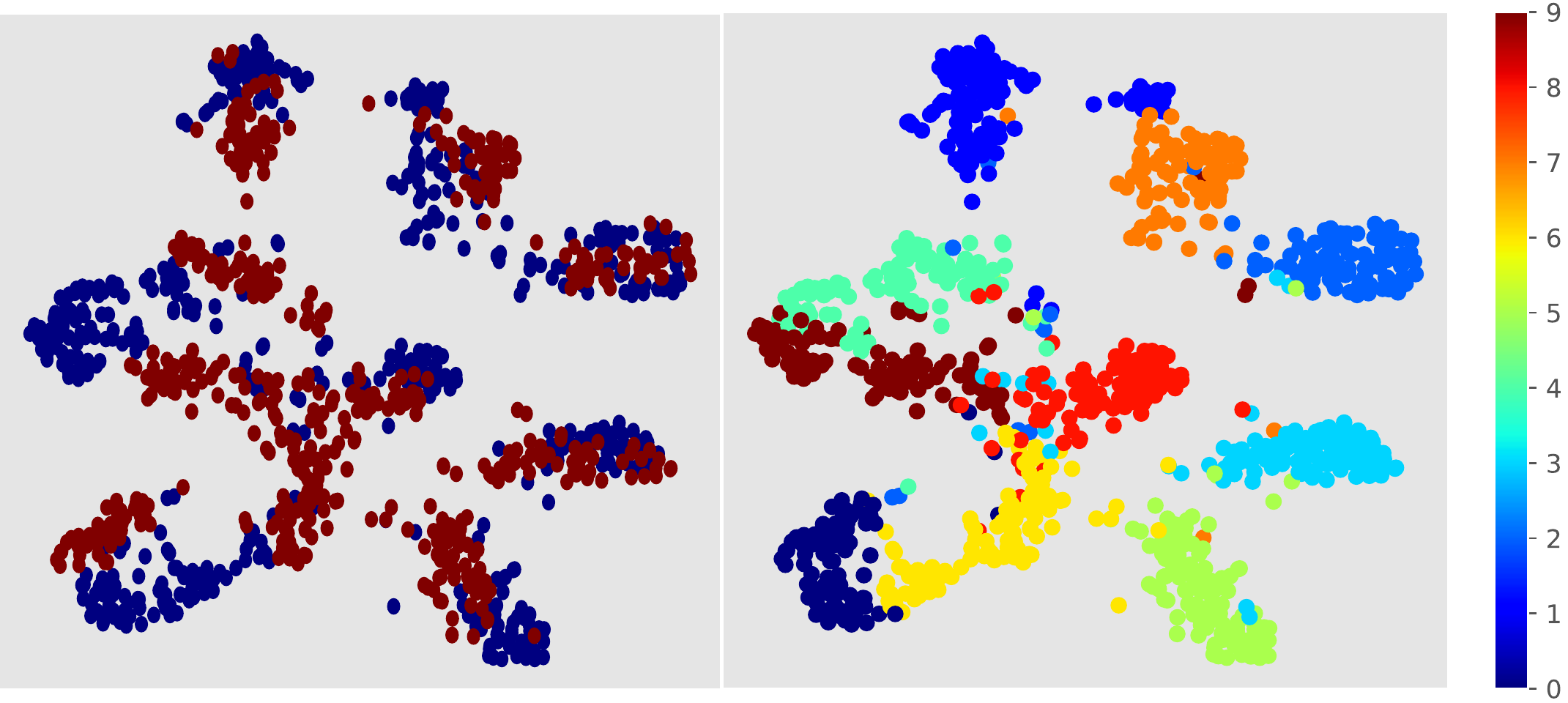}}
	\subfigure[ARN]{
		\label{fig:arn-tsne}
		\includegraphics[width=0.3\textwidth]{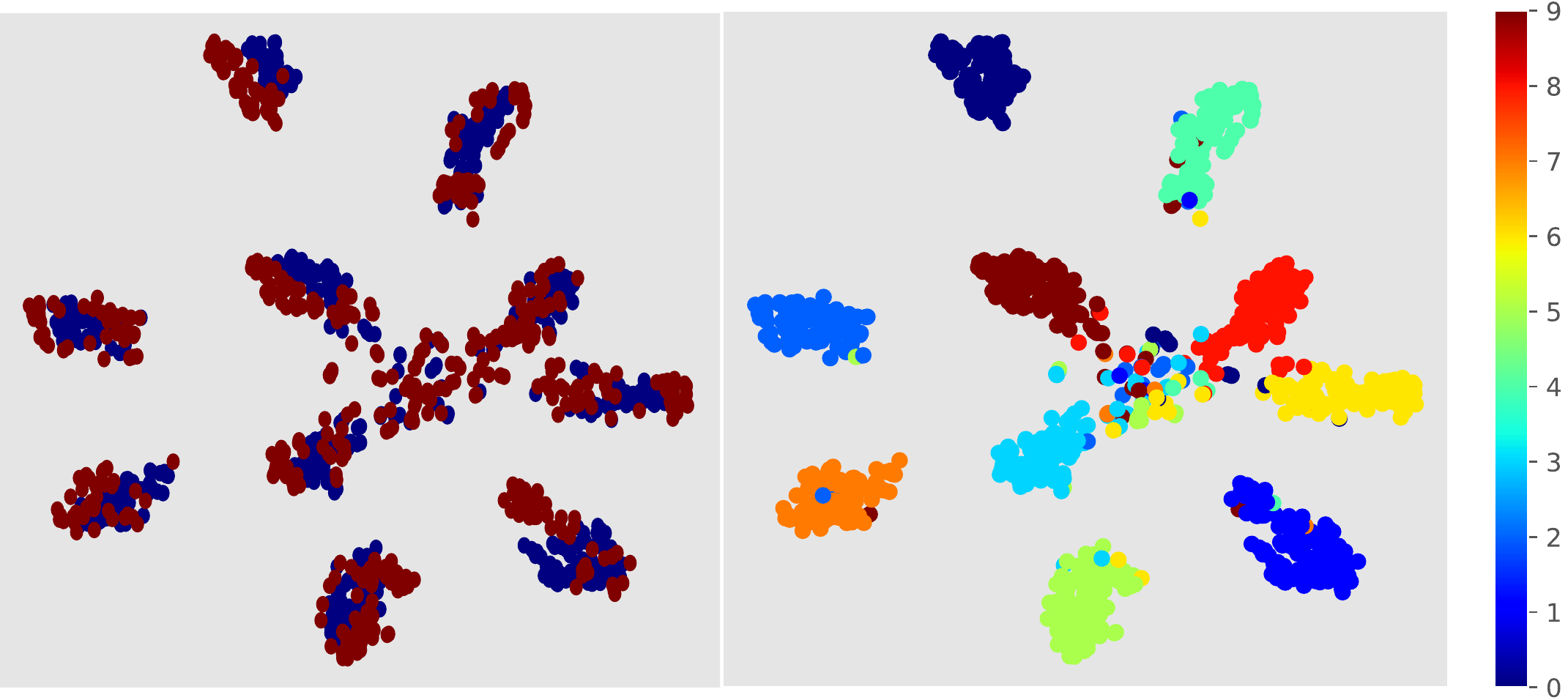}}
	\caption{T-SNE visualization on \textbf{SVHN$\to$MNIST} with their corresponding domain labels (red: target; blue: source) and category labels (10 classes) shown in the left and right subfigures, respectively.}
	\label{fig:visualize-tsne}
\end{figure*}

\textbf{Parameter sensitivity.} We investigate the sensitivity of $\alpha$ and $m$ on \textbf{SVHN$\to$MNIST}. In Table \ref{tab:sensitivity}, the results show that ARN achieves good performance as $\alpha \in [0.01,0.1]$. Even with larger $\alpha$, ARN is able to achieve convergence. In comparison, denoting $\alpha$ as the weight of adversarial domain loss ($\mathcal{L}_d$), the DANN cannot converge when $\alpha>0.2$ due to the imbalanced adversarial game between the overwhelming domain classifier and the feature extractor. For the sensitivity of $m$, the accuracy of ARN exceeds 96.0\% as $m\geq1$. In Section~\ref{sec:da-theory}, as $m\geq1$, the decoder serves as a domain classifier. These analyses validate that ARN is more insensitive to the parameters than that of DANN. Even in the worst cases, ARN can always achieve convergence.

\textbf{T-SNE embeddings.} We analyze the performance of domain alignment for DANN (DAT)~\citep{ganin15} and ARN (MDAT) by plotting T-SNE embeddings of the features $\textbf{z}$ on the task \textbf{SVHN$\to$MNIST}. In Figure \ref{fig:source-tsne}, the source-only model obtains diverse embeddings for each category but the domains are not aligned. In Figure \ref{fig:dann2-tsne}, the DANN aligns two domains but the decision boundaries of the classifier are vague. In Figure \ref{fig:arn-tsne}, the proposed ARN effectively aligns two domains for all categories and the classifier boundaries are much clearer.

\textbf{Interpreting adapted features via reconstruction.} One of the key advantages of ARN is that by visualizing the reconstructed target images we can infer how the features are domain-invariant. We reconstruct the MDAT features of the target domain and visualize them in Table \ref{tab:visual}. It is observed that the target features are reconstructed to source-like images by the decoder $G_r$. As discussed before, intuitively, MDAT forces the target feature to mimic the source feature distribution, which conforms with the visualization. Similar to image-to-image translation, this indicates that our method conducts implicit feature-to-feature translation that transfers the target features to source-like features, and hence the features are domain-invariant.

\section{Conclusion}
We proposed a new domain alignment approach namely Max-margin Domain-adversarial Training (MDAT) and a MDAT-based deep neural network for unsupervised domain adaptation. The proposed method offers effective and stable gradients for feature learning via an adversarial game between the feature extractor and the reconstruction network. The theoretical analysis provides justifications on how it minimizes the distribution discrepancy. Extensive experiments demonstrate the effectiveness of our method and we further interpret the features by visualization that conforms with our insight. Potential evaluation on semi-supervised learning constitutes our future work.

\bibliography{example_paper}
\bibliographystyle{icml2020}


 \clearpage
\section*{Appendix}
\textbf{Hyperparameter} For all tasks, we simply use the same hyperparameters that are chosen from the sensitivity analysis. We use $\alpha=0.02$ and $m=5.0$, and we reckon that better results can be obtained by tuning the hyperparameters for specific tasks.

%

\subsection*{Sensitivity}
\begin{figure}[t]
	\vspace{-5cm}
	\centering  
	\subfigure[Loss penalty $\alpha$]{
		\label{fig:acc-alpha}
		\includegraphics[width=0.48\textwidth]{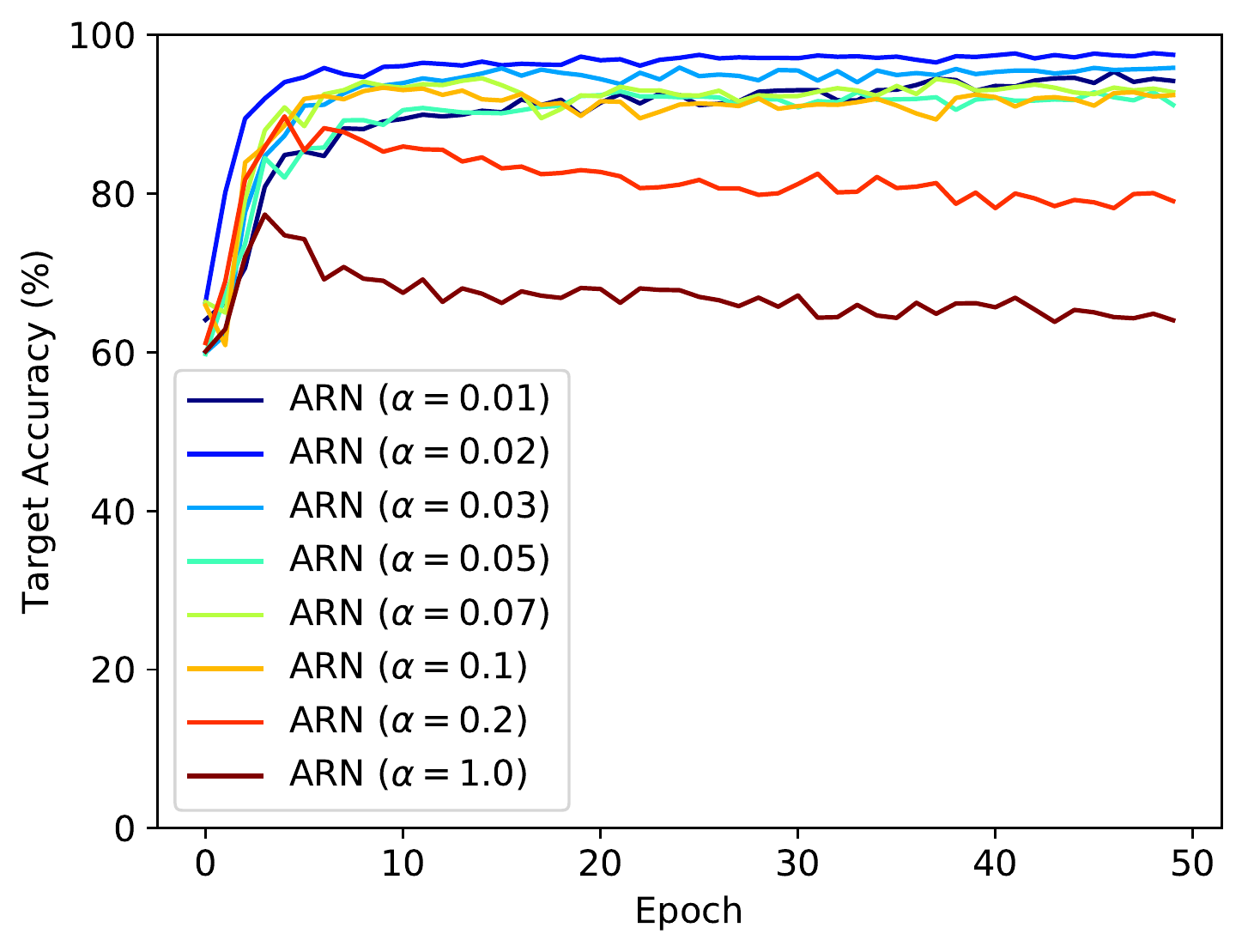}}
	
	\subfigure[Margin $m$]{
		\label{fig:acc-margin}
		\includegraphics[width=0.48\textwidth]{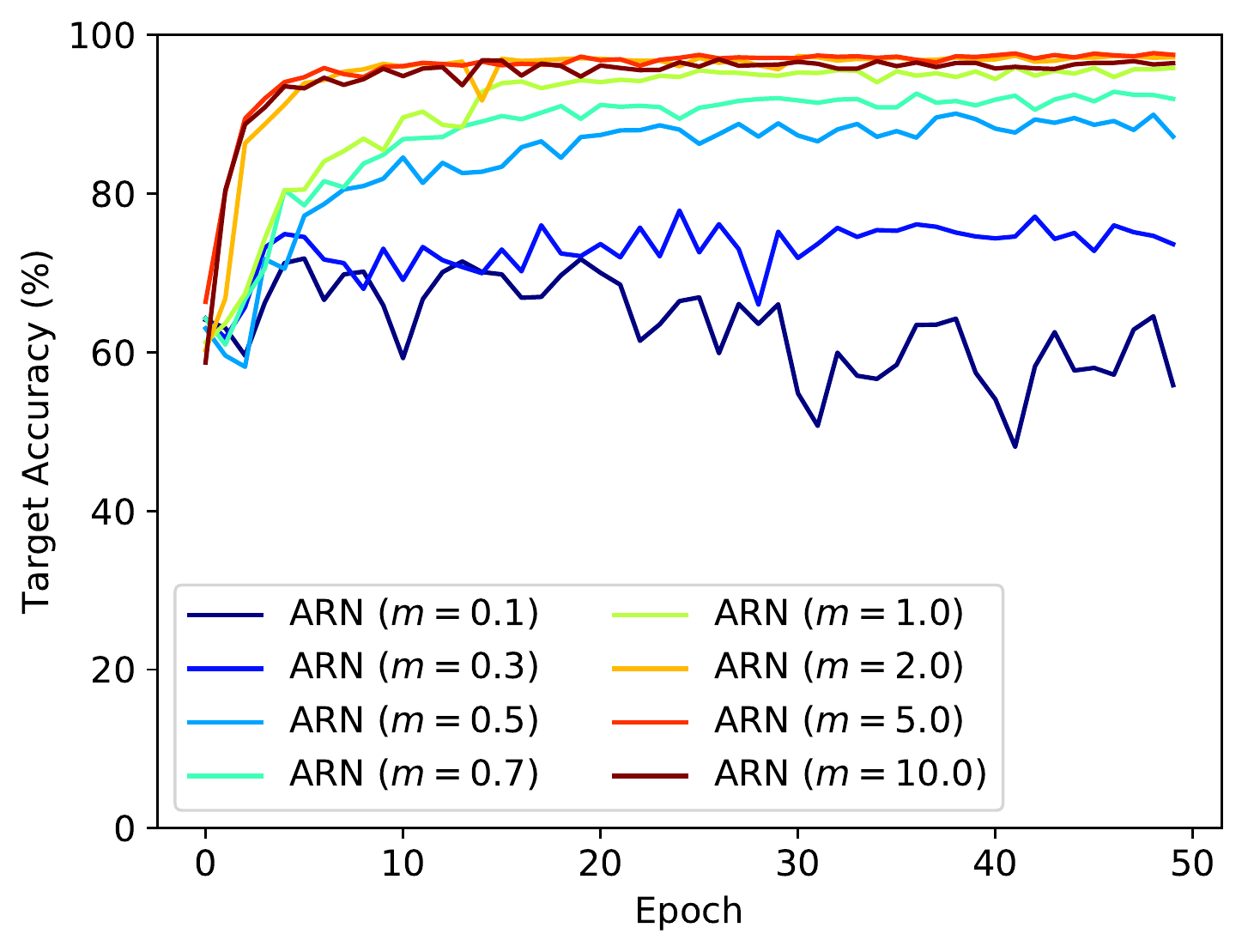}}
	\caption{The training procedure of ARN with different hyper-parameters.}
	\label{fig:more-sentivity}
\end{figure}

We have presented all the results of the sensitivity study in Section \ref{sec:analyses}, and now we show their detailed training procedures in Figure \ref{fig:acc-alpha} and \ref{fig:acc-margin}. It is observed that the accuracy increases when $\alpha$ drops or the margin $m$ increases. The reason is very simple: (1) when $\alpha$ is too large, it affects the effect of supervised training on source domain; (2) when the margin $m$ is small, the divergence between source and target domain (\textit{i.e.} $\mathcal{H}\triangle\mathcal{H}$-distance) cannot be measured well. 

\subsection*{Visualization}
\begin{figure*}[htp]
	\centering 
	\subfigure[\textbf{MNIST$\to$USPS}] {\label{fig:m2u} 
		\includegraphics[width=0.8\columnwidth]{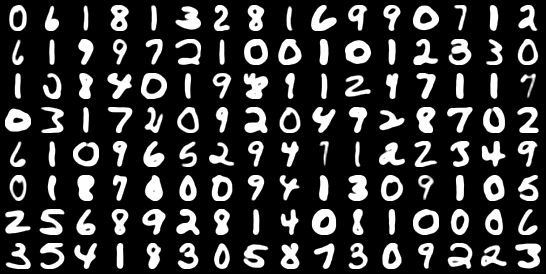}
		\includegraphics[width=0.8\columnwidth]{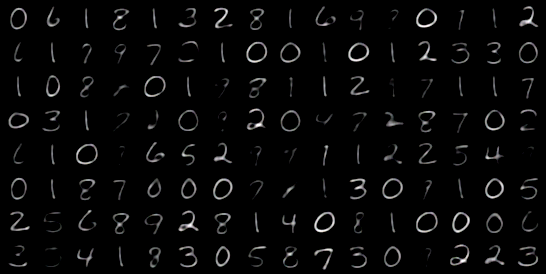}} 
	\subfigure[\textbf{USPS$\to$MNIST}] {\label{fig:u2m} 
		\includegraphics[width=0.8\columnwidth]{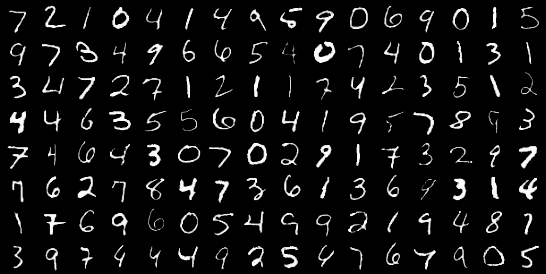}
		\includegraphics[width=0.8\columnwidth]{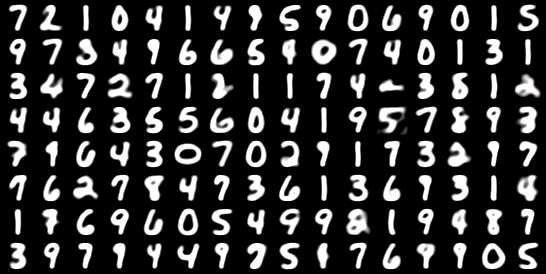}} 
	\subfigure[\textbf{SVHN$\to$MNIST}] {\label{fig:s2m} 
		\includegraphics[width=0.8\columnwidth]{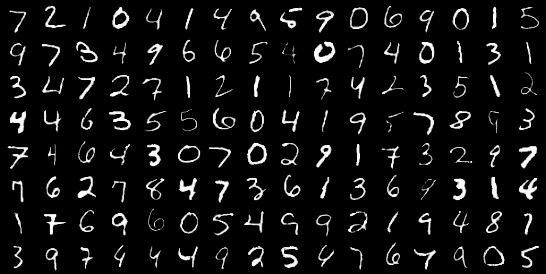}
		\includegraphics[width=0.8\columnwidth]{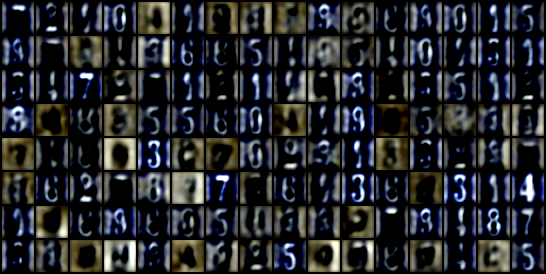}} 
	\subfigure[\textbf{SYN$\to$SVHN}] {\label{fig:s2s} 
		\includegraphics[width=0.8\columnwidth]{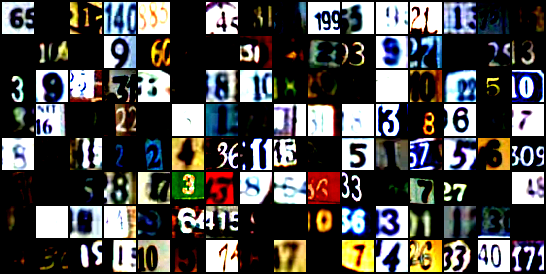}
		\includegraphics[width=0.8\columnwidth]{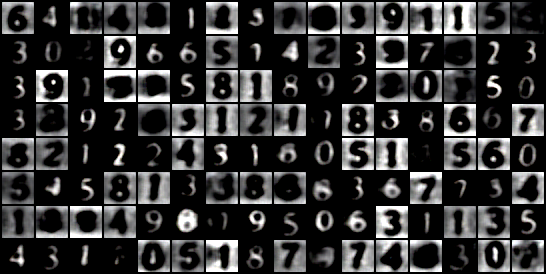}} 
	\caption{Visualization of the target samples and their corresponding reconstructed target samples.} 
	\label{fig:visual_more} 
\end{figure*}

Here we provide more visualization of the reconstructed images of target samples. In Figure \ref{fig:visual_more}, the target samples are shown in the left column while their corresponding reconstructed samples are shown in the right. We can see that for low-level \textit{domain shift} such as \textbf{MNIST$\leftrightarrow$USPS}, the reconstructed target samples are very source-like while preserving their original shapes and skeletons. However, for larger \textit{domain shift} in Figure \ref{fig:s2m} and \ref{fig:s2s}, they are reconstructed to source-like same digits but simultaneously some noises are removed. Specifically, in Figure \ref{fig:s2s}, we can see that one target sample (SVHN) may contain more than one digits that are noises for recognition. After reconstruction, only the right digits are reconstructed. Some target samples may suffer from terrible illumination conditions but their reconstructed digits are very clear, which is amazing.

\end{document}